\documentclass[]{article}
\usepackage{colacl, graphics}

\title{Message Classification in the Call Center}
\author{Stephan Busemann, Sven Schmeier, Roman G.\ Arens\\
DFKI GmbH\\
Stuhlsatzenhausweg 3, D-66123 Saarbr\"{u}cken, Germany\\
\email: {\tt \{busemann, schmeier, arens\}@dfki.de}
}
\date{}

\begin{document}
\newcommand{\email}{e-mail}
\newcommand{\emails}{e-mails}
\newcommand{\smes}{{\sf smes}}
\newcommand{\icc}{{\sc ICC-mail}}

\maketitle

\abstract{Customer care in technical domains is increasingly based on e-mail
communication, allowing for the reproduction of approved
solutions. Identifying the customer's problem is often time-consuming,
as the problem space changes if new products are launched. This paper
describes a new approach to the classification of e-mail requests
based on shallow text processing and machine learning techniques. 
It is implemented within an assistance system for
call center agents that is used in a commercial setting.} 

\section{Introduction}
\label{intro}

Customer care in technical domains is increasingly based on \email\
communication, allowing for the reproduction of approved
solutions. For a call center agent, identifying the customer's problem
is often time-consuming, as the problem space changes if new products
are launched or existing regulations are modified. The typical task of
a call center agent processing \email\ requests consists of the
following steps:
\begin{description}
\item[Recognize the problem(s):] read and understand the \email\ request;
\item[Search a solution:] identify and select predefined text
blocks;
\item[Provide the solution:] if necessary, customize text blocks to meet
the current request, and send the text. 
\end{description}

This task can partly be automated by a system suggesting
relevant solutions for an incoming \email. This would cover the first
two steps. The last step can be delicate, as its primary goal is to keep
the customer satisfied. Thus human intervention seems mandatory to allow for
individual, customized answers. Such a system will 
\begin{itemize}
\item reduce the training effort required since agents don't have to know
every possible solution for every possible problem;
\item increase the agents' performance since agents can more quickly
select a solution among several offered than searching one;
\item improve the quality of responses since agents will behave more
homogeneously -- both as a group and over time -- and commit fewer errors.
\end{itemize}

Given that free text about arbitrary topics must be processed, in-depth
approaches to language understanding are not feasible. Given further
that the topics may change over time, a top-down approach to knowledge
modeling is out of the question. Rather a combination of shallow
text processing (STP) with statistics-based machine learning techniques
(SML) is called for. STP gathers partial information about text such as
part of speech, word stems, negations, or sentence type. These types of
information can be used  to identify the linguistic properties of a
large training set of categorized \emails. SML techniques are used to
build a classifier that is used for new, incoming messages. Obviously,
the change of topics can be accommodated by adding new categories and
\emails\ and producing a new classifier on the basis of old and new
data. We call this replacement of a classifier ``relearning''.

This paper describes a new approach to the classification of \email\ requests
along these lines. It is implemented within the \icc\ system, which is an
assistance system for call center agents that is currently used in a commercial
setting. Section~\ref{data} describes important properties of the input
data, i.e.\ the \email\ texts on the one hand, and the categories on
the other. These properties influenced the system architecture, which
is presented in Section~\ref{lt-ml}. Various publicly available SML
systems  have been tested with different methods of STP-based
preprocessing. Section~\ref{results} describes the results. The implementation
and usage of the system including the graphical user interface is presented in
Section~\ref{impl-use}. We conclude by giving an outlook to further
expected improvements (Section~\ref{conclusions}).

\section{Data Characteristics}
\label{data}
A closer look at the data the \icc\ system is processing will clarify
the task further. We carried out
experiments with unmodified \email\ data 
accumulated over a period of three months in the call center
database. The total amount was 4777 \emails.
We used 47 categories, which contained at least 30
documents. This minimum amount of documents
turned out to render the 
category sufficiently distinguishable for the SML tools. The database contained
74 categories with at least 10 documents, 
but the selected ones covered 94\% of all \emails, i.e.\ 4490 documents.

It has not yet generally been investigated how the
type of data influences the learning result \cite{Yang-99},
or under which circumstances which kind of preprocessing and which
learning algorithm is most appropriate. Several aspects must be considered:
Length of the documents, morphological and syntactic well-formedness,
the degree to which a document can be uniquely classified,
and, of course, the language of the documents. 

In our application 
domain the documents differ very much from documents generally used in
benchmark tests, for example the Reuters corpus\footnote{{\tt
http://www.research.att.com/\~{}lewis/reuters21578. html}}. First of all, 
we have to deal with German, whereas the Reuters data are in English.
The average length of our \emails\ is 60 words,
whereas for documents of Reuters-21578 it is 129 words. 
The number of categories we used compares to the top 47 categories of
the  Reuters {\it TOPICS} category set. While we have 5008 documents,
{\it TOPICS} consists of 13321 instances\footnote{We took only uniquely
classified documents into account.}.  
The Reuters documents usually are morphologically and syntactically
well-formed. As \emails\ are a more spontaneously created and informal type of
document, they require us to cope with  a
large amount of jargon, misspellings and grammatical
inaccuracy. A drastic example is shown in Figure~\ref{Figure3}.
The bad conformance to linguistic standards was a major argument in
favor of STP instead of in-depth syntactic and semantic analysis.

The degree to which a document can be uniquely classified
is hard to verify and can only be inferred from the results in
general terms.\footnote{Documents containing
multiple requests can at present only be treated manually, as 
described in Section~\ref{impl-use}.} 
It is, however, dependent on the ability to uniquely
distinguish the classes. In our application we encounter overlapping
and non-exhaustive categories as the category system develops over
time.

\section{Integrating Language Technology With Machine Learning}
\label{lt-ml}
STP and SML correspond to two different paradigms. STP tools used for
classification tasks promise very high recall/precision or accuracy
values. Usually human experts define one or several template structures
to be filled automatically by extracting information from the 
documents (cf.\ e.g.\ \cite{FACILE-99}). Afterwards, the partially
filled templates are classified by hand-made rules. The whole process
brings about high costs in analyzing 
and modeling the application domain, especially if it is to take into
account the problem of changing categories in the present application. 

SML promises low costs both in analyzing and modeling the 
application at the expense of a lower accuracy. It
is independent of the domain on the one hand, but does not consider
any domain specific knowledge on the other. 

By combining both methodologies in \icc, we achieve high accuracy and can still
preserve a useful degree of domain-independence. STP may use both general
linguistic knowledge and linguistic algorithms or heuristics adapted
to the application in order to extract information from texts that is relevant 
for classification. The input to the SML tool is enriched with that
information. The tool builds one or several categorizers\footnote{Almost
all tools we examined build a single multi-categorizer except for SVM\_Light,
which builds multiple binary classifiers.} that will classify new
texts. 

In general, SML tools work with a vector representation of
data. First, a {\em relevancy vector\/} of relevant features for each class
is computed \cite{Yang-97}. In our case the relevant
features consist of the user-defined output of the linguistic
preprocessor. Then  
each single document is translated into a vector of numbers isomorphic
to the defining vector. Each entry represents the occurrence of the
corresponding feature. More details will be given in
Section~\ref{results}

\begin{figure*}
\begin{center}
\includegraphics{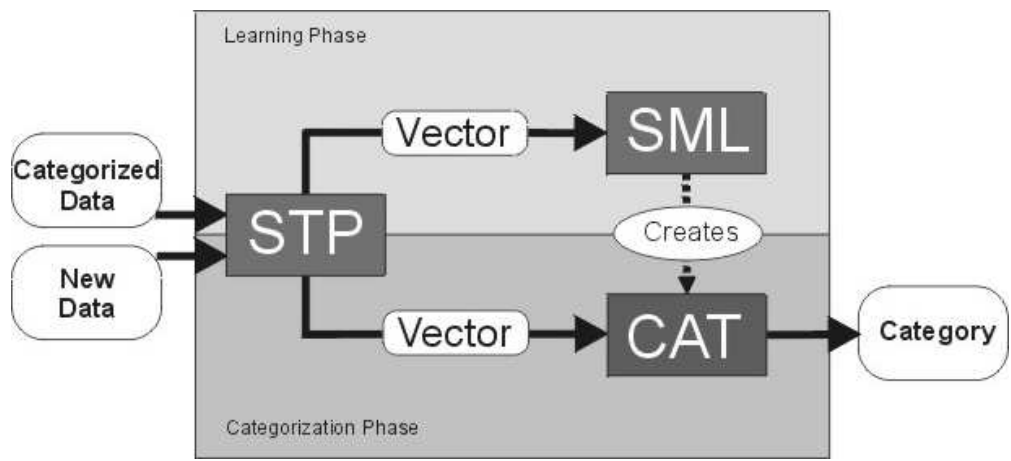}
\end{center}
\caption{Architecture of the \protect\icc\ System. \label{Figure1}}
\end{figure*}

The \icc\ architecture is shown in Figure~\ref{Figure1}. The
workflow of the system consists of a learning step carried out off-line
(the light gray box) and an online categorization step (the dark gray box). In
the off-line part, categorizers are built by processing classified
data first by an STP and then by an SML tool. In this way, categorizers can be
replaced by the system administrator as she wants to include new or
remove expired categories. The categorizers are used on-line in order to
classify new documents after they have passed the linguistic
preprocessing. The resulting category is in our application associated
with a standard text that the call center agent uses in her answer. The
on-line step provides new classified data that is stored in a
dedicated \icc\ database (not shown in Figure~\ref{Figure1}). The
relearning step is based on data from this database.

\subsection{Shallow Text Processing}
Linguistic preprocessing of text documents is carried out by re-using \smes, 
an information extraction core system for real-world German text processing 
\cite{Neumann-97}. The fundamental design criterion of \smes\ is to provide 
a set of basic, powerful, robust, and efficient STP components and 
generic linguistic knowledge sources that can easily be customized to 
deal with different tasks in a flexible manner. \smes\
includes a text tokenizer, a lexical processor and a chunk parser. 
The chunk parser itself is subdivided into three components. In
the first step, phrasal fragments  like general nominal 
expressions and verb groups are recognized. Next,
the dependency-based structure of 
the fragments of each sentence is computed using a set of specific sentence 
patterns. Third, the grammatical functions are determined for each
dependency-based structure on the basis of a large subcategorization
lexicon. 
The present application benefits from the high modularity of the usage
of the components. Thus, it is possible to run only a subset of 
the components and to tailor their output. The experiments described in
Section~\ref{results} make use of this feature.


\subsection{Statistics-Based Machine Learning}
Several SML tools representing different learning paradigms
have been selected and evaluated in different settings of our domain:
\begin{description}
\item[Lazy Learning:] Lazy Learners
are also known as memory-based, instance-based, exemplar-based,
case-based, experience-based, or $k$-nearest neighbor 
algorithms. They store all documents as vectors during the learning phase. In
the categorization phase, the new document vector is compared to the stored
ones and is categorized to same class as the $k$-nearest neighbors. The
distance is measured by computing e.g.\ 
the Euclidean distance between the vectors.
By changing the number of neighbors $k$ or the kind of distance measure,
the amount of generalization
can be controlled.

We used IB \cite{Aha-92}, which is part of the MLC++ library
\cite{Kohavi-96}.

\item[Symbolic Eager Learning:] 
This type of learners constructs a representation for document vectors
belonging to 
a certain class during the learning phase, e.g.\ decision trees,
decision rules or probability weightings. During the categorization
phase, the representation is used to assign the appropriate class to a
new document vector. Several pruning or specialization heuristics can
be used to control the amount of generalization. 

We used ID3 \cite{Quinlan-86}, C4.5
\cite{Quinlan-92} and C5.0,  RIPPER \cite{Cohen-95}, and the Naive Bayes
inducer \cite{Good-65} contained
in the  MLC++ library. ID3, C4.5 and C5.0 produce decision trees,
RIPPER is a rule-based learner and the Naive Bayes algorithm computes
conditional probabilities of the classes from the instances.

\item[Support Vector Machines (SVMs):]
SVMs are described in \cite{Vapnik-95}. SVMs are binary learners in
that they distinguish positive and negative examples for each
class. Like eager learners, they construct a representation during the
learning phase, namely a hyper plane supported by vectors of positive
and negative examples. For each class, a  categorizer is built by
computing such a hyper plane.
During the categorization phase, each categorizer is applied to the new
document vector, yielding the probabilities of the document belonging to a
class. The probability increases with the distance of
thevector from the hyper plane. A document is said to belong to the
class with the highest probability. 

We chose SVM\_Light \cite{Joachims-98}. 

\item[Neural Networks:]
Neural Networks are a special kind of ``non-symbolic'' eager learning
algorithm. The neural network links the vector elements to the
document categories The learning phase defines
thresholds for the activation of neurons. In the categorization phase,
a new document vector leads to the activation of a single category. For
details we refer to \cite{Wiener-95}.  

In our application, we tried out the
Learning Vector Quantization (LVQ) \cite{Kohonen-96}. LVQ has been used
in its default configuration only. 
No adaptation to the application domain has been made.
\end{description}

\section{Experiments and Results}
\label{results}
We describe the experiments and results we achieved with different linguistic
preprocessing and learning algorithms and provide some interpretations. 

We start out from the corpus of categorized \emails\ described in
Section~\ref{data}. In order to normalize the vectors representing the
preprocessing results of texts of different length, and to concentrate
on relevant material (cf.\
\cite{Yang-97}), we define the relevancy vector as follows. First, all 
documents are  preprocessed, yielding a 
list of results for each category. From each of these lists, the
100 most  frequent results -- according to a TF/IDF measure -- are
selected. The relevancy vector consists of all selected results, where
doubles are eliminated. Its length was about 2500 for the 47
categories; it slightly varied with the kind of preprocessing used.

During the learning phase, each document is preprocessed. The result
is mapped onto a vector of the same length as the relevancy vector. For every
position in the relevancy vector, it is determined whether the
corresponding result has been found. In that case, the value of the
result vector element is 1, otherwise it is 0.

In the categorization phase, the new document is preprocessed, and a
result vector is built as described above and handed over to the
categorizer (cf.\ Figure~\ref{Figure1}). 

While we tried various kinds of linguistic preprocessing, systematic
experiments have been carried out with morphological analysis ({\em
MorphAna\/}), shallow parsing heuristics ({\em STP-Heuristics\/}), and a
combination of both ({\em Combined\/}).

\begin{description}
\item[MorphAna:] Morphological Analysis provided by \smes\ yields the
word stems of nouns, verbs and adjectives, as well as the full forms of
unknown  words. We are using a lexicon of approx.\ 100000
word stems of German \cite{Neumann-97}.

\item[STP-Heuristics:] Shallow parsing techniques are used to
heuristically identify sentences
containing relevant information. The \emails\ usually contain questions and/or
descriptions of problems. The manual analysis of a sample of the data suggested
some linguistic constructions frequently used to express the problem.
We expected that content words in these constructions should be particularly
influential to the categorization.  Words in these
constructions are extracted and processed as in {\em MorphAna}, and all other
words are ignored.\footnote{If no results were found this way, {\em
MorphAna\/} was 
applied instead.} The heuristics were implemented in \icc\ using \smes.

The constructions of interest include negations at the sentence and
the phrasal level, yes-no and wh-questions, and declaratives
immediately preceding questions. Negations were found to
describe a state to be changed or to refer to missing objects,
as in {\em I cannot read my email\/} or {\em There is no correct
date}. We identified them through negation particles.\footnote{We
certainly would have benefited from lexical semantic information, e.g.\
{\em The correct date is missing\/} would not be captured by our approach.}
Questions most often refer to the problem in hand, either directly,
e.g.\ {\em How can I start my email 
program?\/} or indirectly, e.g.\ {\em Why is this the case?}. The
latter most likely refers to the preceding sentence, e.g.\ {\em My
system drops my e-mails.} Questions are identified by their word order,
i.e.\ yes-no questions start with a verb and wh-questions with a
wh-particle. 

\item[Combined:] In order to
emphasize words found relevant by the STP heuristics without losing
other information retrieved by {\em MorphAna}, the previous two
techniques are combined. Emphasis is represented here
by doubling the number of occurrences of the tokens in the
normalization phase, thus increasing their TF/IDF value.
\end{description}

Call center agents judge the performance of \icc\ most easily in terms
of accuracy: 
In what percentage of cases does the classifier suggest the
correct text block? In Table~\ref{Figure2}, detailed information about
the accuracy achieved is presented. All experiments were carried
out using 10-fold cross-validation on the data described in
Section~\ref{data}. 
\begin{table*}

\begin{center}
\begin{tabular}{|l||l|lr|lr|lr|}
\hline
&&MorphAna && STP-Heuristics && Combined& \\
&{\bf SML algorithm} &Best& Best5&Best& Best5&Best&Best5\\
\hline
\hline
Neural Nets&{\bf LVQ} & 35.66 && 22.29 && 25.97&\\
\hline
\hline
Lazy Learner&{\bf IB} & 33.81 && 33.01 && 35.14&\\
\hline
\hline
Symbolic Eager&{\bf Naive Bayes} & 33.83 && 33.76 && 34.01&\\
\hline
Learners&{\bf ID3} & 38.53 && 38.11 && 40.02&\\
\hline
&{\bf RIPPER} & 47.08 && 49.38 && 50.54&\\
\hline
&{\bf Boosted Ripper} & 52.73 && 49.96 && 50.78&\\
\hline
&{\bf C4.5} & 52.00 && 52.90 && 53.40&\\
\hline
&{\bf C5.0} & 52.60 && 53.20 && 54.20&\\
\hline
\hline
Support Vectors&{\bf SVM\_Light} & 53.85 & 74.91 & 54.84 & 78.05 & 56.23 & 78.17\\

\hline
\end{tabular}
\caption{Results of Experiments. Most SML tools deliver the best result
only. SVM\_Light produces ranked results, allowing to measure the
accuracy of the top five alternatives (Best5).\label{Figure2}} 
\end{center}
\end{table*}

In all experiments the SVM\_Light system outperformed other learning
algorithms, which confirms Yang's \cite{Yang-99b} results for SVMs fed
with Reuters data. The 
$k$-nearest neighbor algorithm IB performed surprisingly badly although 
different values of $k$ were used. For IB, ID3, C4.5,
C5.0, Naive Bayes, RIPPER and SVM\_Light, linguistic 
preprocessing increased the overall performance. In fact, the method
performing best,
SVM\_Light, gained 3.5\% by including the task-oriented heuristics.
However, the boosted RIPPER and LVQ scored a decreased accuracy value there.
For LVQ the decrease may be due to the fact that no adaptations to the domain
were made, such as adapting the number of codebook vectors, the
initial learning parameters or the number of iterations during training
(cf.\ \cite{Kohonen-96}). Neural 
networks are rather sensitive to misconfigurations.
The boosting for RIPPER seems to run into problems of overfitting. We noted
that in six trials the accuracy could be improved in {\em Combined\/}
compared to {\em MorphAna}, but in four trials, boosting led to
deterioration. This effect is also mentioned in \cite{Quinlan-96}.

These figures are slightly lower than the ones reported by
\cite{Schmeier-99} that were obtained from a different data
set. Moreover, these data did not contain multiple queries in
one \email.

It would be desirable to provide explanations for the behavior of the
SML algorithms on our data. As we have emphasized in
Section~\ref{data}, general methods of explanation do not exist yet. In
the application in hand, we found it difficult to account for the
effects of e.g.\ ungrammatical text or redundant categories. For the
time being, we can only
offer some speculative and inconclusive assumptions: Some of the tools
performing badly 
-- IB, ID3, and the Naive Bayes inducer of the MLC++ library -- have no
or little pruning ability. With rarely occurring data, this leads to
very low generalization rates, which again is
a problem of overfitting. This suggests that a more canonical
representation for the many ways of expressing a technical problem
should be sought for. Would more extensive linguistic preprocessing help?

Other tests not reported in Table~\ref{Figure2} looked at improvements through
more general and sophisticated STP such as chunk parsing. The results
were very discouraging, 
leading to a significant decrease compared to {\em MorphAna}. We
explain this with the bad compliance of \email\
texts to grammatical standards (cf.\ the example in Figure~\ref{Figure3}). 

However, the practical usefulness of chunk
parsing or even deeper language understanding such as semantic analysis
may be questioned in general: In a moving domain, the coverage of linguistic
knowledge will always be incomplete, as it 
would be too expensive for a call center to have language technology
experts keep pace with the occurrence of new topics. Thus the
preprocessing results will  often differ for \emails\ expressing the
same problem and hence not be useful for SML.  

As a result of the tests in our application domain,  we
identified a favorite statistical tool and found that task-specific linguistic
preprocessing is encouraging, while general STP is not.

\section{Implementation and Use}
\label{impl-use}
\begin{figure*}
\begin{center}
\includegraphics{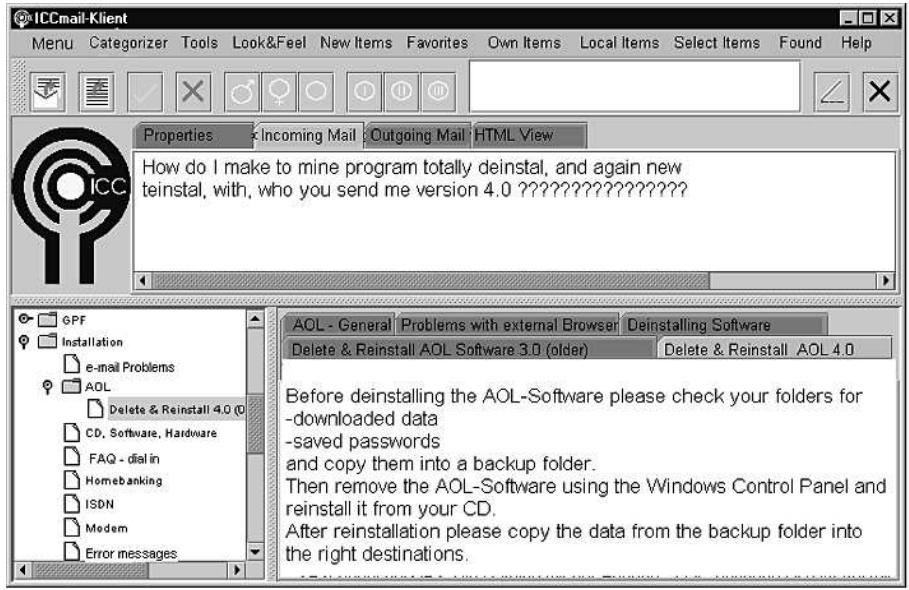}
\end{center}
\caption{The GUI of the \protect\icc\ Client. All labels and texts were
translated by the authors. The English input is based on the following
original text, which is similarly awkward though understandable: {\em 
Wie mache ich zum mein Programm total deinstalieren, und wieder neu   
instalierem, mit, wen Sie mir senden Version 4.0 ??????????????}. The
suggested answer text is associated with the category named ``Delete~\&
Reinstall AOL~4.0''. Four alternative answers can be selected using the
tabs. The left-hand side window displays the active category in context.}
\label{Figure3}
\end{figure*}

In this section we describe the integration of the \icc\ system into the
workflow of the call center of AOL Bertelsmann Online GmbH~\&~Co.\ KG,
which answers requests about the German version of AOL
software. A client/server solution was built that allows the call
center agents to connect as clients to the \icc\ server, which implements the
system described in Section~\ref{lt-ml}. For this purpose, it was
necessary to
\begin{itemize}
\item connect the server module to AOL's own Sybase database that
delivers the incoming mail and dispatches the outgoing answers, and to
\icc's own database that stores the classified \email\ texts;
\item design the GUI of the client module in a self-explanatory and
easy to use way (cf.\ Figure~\ref{Figure3}). 
\end {itemize}

The agent reads in an \email\ and starts \icc\ using GUI buttons. She verifies
the correctness of the suggested answer, displaying and perhaps selecting
alternative solutions. If the agent finds the appropriate answer within
these proposals, the associated text is filled in at the correct
position of the answer \email. If, on the other hand, no proposed
solution is found to be adequate, the \icc\ tool can still be used to
manually select any text block from the database. The \icc\ client had
to provide the functionality of the tool already in use since an
additional tool was not acceptable to the agents, who are working under
time pressure. 

In the answer \email\ window, the original \email\ is automatically
added as a quote. If an \email\ contains several questions, the
classification process
can be repeated by marking each question and iteratively 
applying the process to the marked part. The agent can edit the
suggested texts before sending them off. 
In each case, the classified text together with the selected category is
stored in the \icc\ database for use in future learning steps. 

Other features of the \icc\ client module include a spell checker
and a history view. The latter displays not only the previous \emails\ of
the same  author but also the solutions that have been proposed and the
elapsed time before an answer was sent. 

The assumed average time for an agent to answer an \email\ is a bit
more than two minutes
with AOL's own mail processing system.\footnote{This system does not include
automatic analysis of mails.} With the \icc\ system the complete cycle of
fetching the mail, checking the proposed solutions, choosing the
appropriate solutions, inserting additional text fragments and sending
the answer back can probably be achieved in half the time. Systematic
tests supporting this claim are not completed yet,\footnote{As of end of
February 2000.} but the following
preliminary results are encouraging: 
\begin{itemize}
\item
A test under real-time
conditions at the call-center envisaged the use of the
\icc\ system as a mail tool only, i.e.\
without taking advantage of the system's intelligence. It showed that
the surface and the look-and-feel is accepted and the functionality
corresponds to the real-time needs of the call center agents, as users were
slightly faster than within their usual environment. 
\item
A preliminary test of the throughput achieved by using the STP and SML
technology in \icc\ showed that experienced users take about 50-70
seconds on average for one cycle, as described above.
This figure was gained through experiments with 
three users over a duration of about one hour each. 
\end{itemize}

Using the system with a constant set of categories will improve its
accuracy after repeating the off-line learning step. If a new category
is introduced, the accuracy will slightly decline until 30 documents
are manually classified and the category is automatically included into
a new classifier. Relearning may take place at regular intervals. The
definition of new categories must be fed into \icc\ by a ``knowledge
engineer'', who maintains the system. The effects of new categories and
new data have not been tested yet.

The optimum performance of \icc\ can be achieved only with a
well-maintained category 
system. For a call center, this may be a difficult task to achieve, espescially
under severe time pressure, but it will pay off. In 
particular, all new categories should be added, outdated ones should be
removed, and redundant ones merged. Agents should only use these
categories and no others. The organizational structure of the team
should reflect this by defining the tasks of the
``knowledge engineer'' and her interactions with the agents.

\section{Conclusions and Future Work}
\label{conclusions}

We have presented new combinations of STP and SML methods to classify
unrestricted \email\ text according to a changing set of categories. 
The current accuracy of the \icc\ system is 78\% (correct solution
among the top
five proposals), corresponding to an overall performance  of 73\%  since
\icc\ processes only 94\% of the incoming \emails. The
accuracy improves with  usage, since each relearning step
will yield better classifiers. The accuracy is expected to approximate that
of the agents, but not improve on it. With \icc, the performance of an
experienced agent can approximately be doubled.

The system is currently undergoing extensive tests at the call center
of AOL Bertelsmann Online. Details about the development of the
performance depending on the throughput and change 
of categories are expected to be available by mid 2000.

Technically, we expect improvements from the following areas of future
work. 
\begin{itemize}
\item Further task-specific heuristics aiming at general 
structural linguistic properties should be defined. This includes
heuristics for the identification of multiple requests in a single
\email\ that could be based on key words and key phrases as well as on
the analysis of the document structure. 
\item Our initial experiments
with the integration of GermaNet \cite{Hamp-97}, the evolving German
version of WordNet, seem to confirm the positive results described for 
WordNet \cite{Rodriguez-97} and will thus be extended. 
\item A  reorganization of the existing three-level
category system into a semantically consistent tree structure would
allow us to explore the non-terminal nodes of the tree for
multi-layered SML. This places additional requirements on the knowledge
engineering task and thus needs to be thoroughly investigated for
pay-off. 
\item Where system-generated answers are acceptable to customers, a
straightforward extension of \icc\ can provide this functionality. For
the application in hand, this was not the case.
\end{itemize}

The potential of the technology presented extends beyond call center
applications. We intend to explore its use within an information broking
assistant in document classification. In a further industrial project
with German Telekom, the \icc\ technology will be extended to process
multi-lingual press releases. The nature of these documents will allow us 
to explore the application of more sophisticated language technologies 
during linguistic preprocessing.

\section*{Acknowledgments}
We are grateful to  our colleagues G\"{u}nter Neumann, Matthias
Fischmann, Volker Morbach, and Matthias Rinck for fruitful discussions
and for support with \smes\ modules. This work was partially
supported by a grant of the Minister of Economy
and Commerce of the Saarland, Germany, to the project ICC.

\bibliographystyle{acl}


\end{document}